%% file: main.tex
\documentclass[10pt,twocolumn,letterpaper]{article}

\usepackage[pagenumbers]{cvpr} 

\usepackage{algorithm}
\usepackage{algorithmic}

\usepackage{amsmath}
\usepackage{amssymb}
\usepackage{booktabs}
\usepackage{multirow} 

\newcommand{\methodname}{DropAnSH-GS}

\input{preamble}
\definecolor{cvprblue}{rgb}{0.21,0.49,0.74}
\usepackage[pagebackref,breaklinks,colorlinks,allcolors=cvprblue]{hyperref}

\def\paperID{44979} 
\def\confName{CVPR}
\def\confYear{2026}

\title{Dropping Anchor and Spherical Harmonics for Sparse-view \\Gaussian Splatting}

\author{Shuangkang Fang\textsuperscript{1}, 
~I-Chao Shen\textsuperscript{2}, 
~Xuanyang Zhang\textsuperscript{3},
~Zesheng Wang\textsuperscript{1},
~Yufeng Wang\textsuperscript{1},\\
~Wenrui Ding\textsuperscript{1}, 
~Gang Yu\textsuperscript{3},
~Takeo Igarashi\textsuperscript{2}\\
\textsuperscript{1}Beihang University~~~ \textsuperscript{2}The University of Tokyo~~~
\textsuperscript{3}StepFun\\
{\textcolor{black}{\tt\small https://sk-fun.fun/DropAnSH-GS}}
}

\begin{document}
\maketitle
\input{sec/0_abstract}    
\input{sec/1_intro}

\input{sec/2_related_work}

\input{sec/3_method}

\input{sec/4_exp}

\input{sec/5_conclusion}

{
    \small
    \bibliographystyle{ieeenat_fullname}
    \bibliography{main}
}

\input{sec/X_suppl}

\end{document}

%% file: sec/0_abstract.tex
\begin{abstract}
Recent 3D Gaussian Splatting (3DGS) Dropout methods address overfitting under sparse-view conditions by randomly nullifying Gaussian opacities.
However, we identify a neighbor compensation effect in these approaches: dropped Gaussians are often compensated by their neighbors, weakening the intended regularization. Moreover, these methods overlook the contribution of high-degree spherical harmonic coefficients (SH) to overfitting.
To address these issues, we propose \methodname{}, a novel anchor-based Dropout strategy. Rather than dropping Gaussians independently, our method randomly selects certain Gaussians as anchors and simultaneously removes their spatial neighbors. This effectively disrupts local redundancies near anchors and encourages the model to learn more robust, globally informed representations.
Furthermore, we extend the Dropout to color attributes by randomly dropping higher-degree SH to concentrate appearance information in lower-degree SH. This strategy further mitigates overfitting and enables flexible post-training model compression via SH truncation.
Experimental results demonstrate that \methodname{} substantially outperforms existing Dropout methods with negligible computational overhead, and can be readily integrated into various 3DGS variants to enhance their performances.
\end{abstract}

%% file: sec/1_intro.tex
\section{Introduction}
Realistic novel view synthesis (NVS) remains a central challenge in computer vision and graphics. 3DGS~\cite{kerbl2023-3dgs}, with a remarkable balance between rendering speed and visual fidelity, has recently emerged as the leading method in this domain~\cite{chen2024survey-3dgs,samavati20233D-rec-survey,fang2024-ce3d,matsuki2024gaussian,jiang2024gaussianshader}. Although 3DGS excels in dense input views~\cite{bag20243dgs-comp-survey, lu2024scaffold,chen2025hac,yu2024mip-gs, liu2024comp-gs, huang20242dgs}, it faces significant challenges when trained with sparse views~\cite{mihajlovic2025splatfields, colmapfree3dgs, fan2024instantsplat, zhang2024corgs,han2024binocular,fang2025nerf-gs}. The scarcity of training views often induces severe overfitting, manifesting as artifacts, blurring, or geometric distortions that limit its practical applicability.

\begin{figure}[t]
\centering
\includegraphics[width=1.\columnwidth]{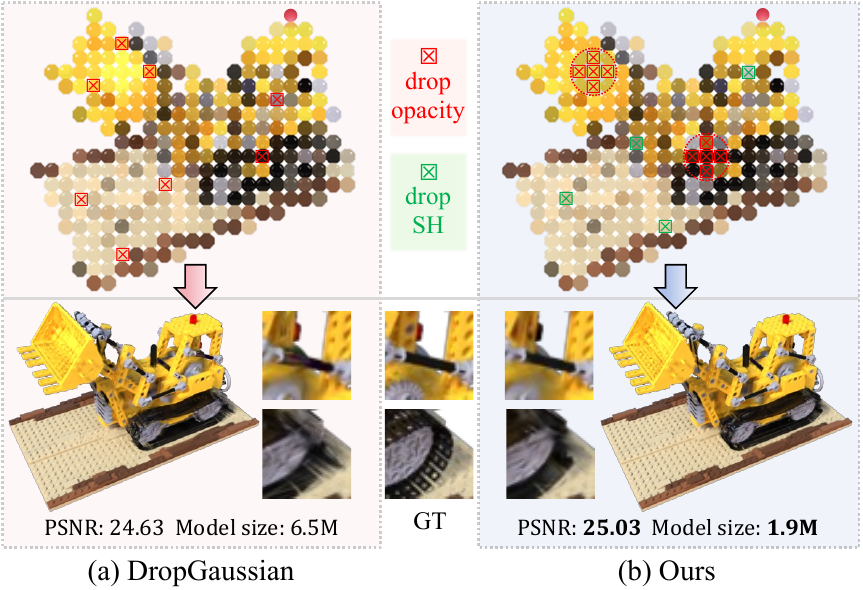}
\caption{\textbf{Difference between DropGaussian~\cite{park2025dropgaussian} and our \methodname{}}. (a) DropGaussian randomly sets the opacity of individual Gaussians to zero. (b) \methodname{} selects random Gaussians as anchors, discards their neighbors, and applies Dropout to high-degree SH. 
Anchor-based Dropout eliminates contiguous 3D regions, which effectively inhibits adjacent Gaussians from compensating for the dropped ones, forcing remaining Gaussians to learn more robust scene representations, thereby strengthening the Dropout regularization effect.
Dropping SH further reduces overfitting and yields a more compact model.
}
\label{fig:teaser}
\end{figure}

\begin{figure*}[t]
\centering
\includegraphics[width=1. \textwidth]{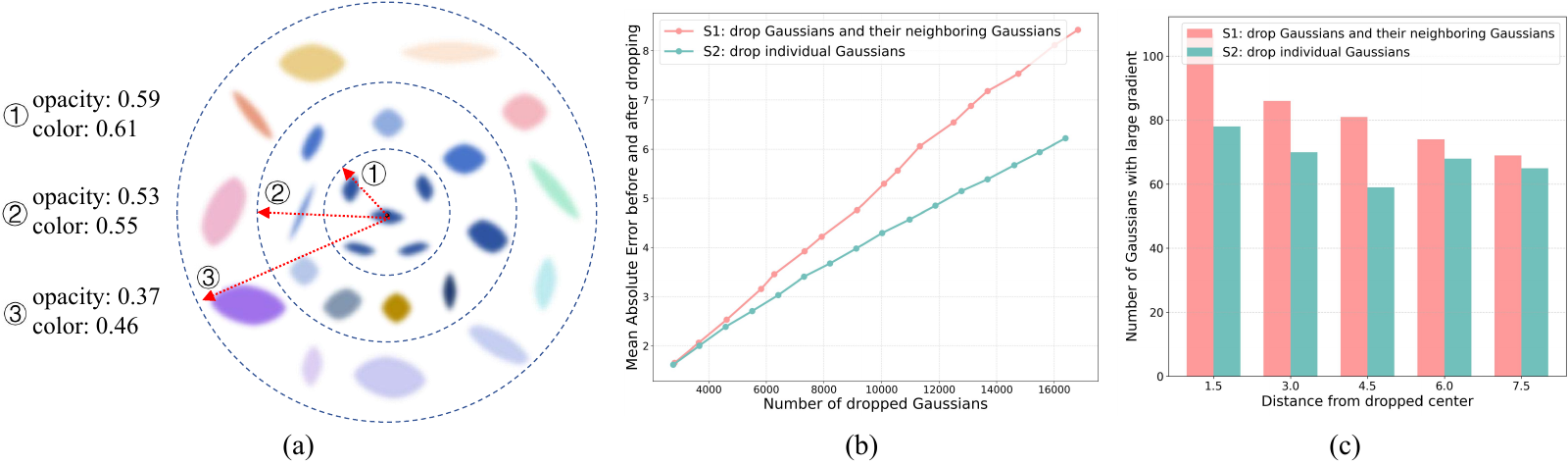} 
\caption{\textbf{Compensation effect in neighbor Gaussians}. (a) We measure the spatial autocorrelation of opacity and color between Gaussians at varying distances using Moran’s I metric~\cite{moran1950notes,mihajlovic2025splatfields}. Closer Gaussians exhibit higher similarity in opacity and color. This spatial redundancy implies that dropping only individual Gaussians can be easily compensated for by their surrounding similar neighbors, thereby limiting the effectiveness of dropout.
(b) We report the Mean Absolute Error between rendered images before and after applying two different Dropout strategies, S1 and S2. When dropping a similar number of Gaussians, the strategy that drops Gaussians and their neighboring Gaussians (S1) has a larger impact on rendering quality, as it avoids simple compensation by neighboring Gaussians.
(c) In regions surrounding the dropped Gaussians, the S1 strategy activates a greater number of remaining Gaussians, resulting in stronger gradient updates.}
\label{fig:pilot_study}
\end{figure*}

\begin{figure}[t]
\centering
\includegraphics[width=0.45 \textwidth]{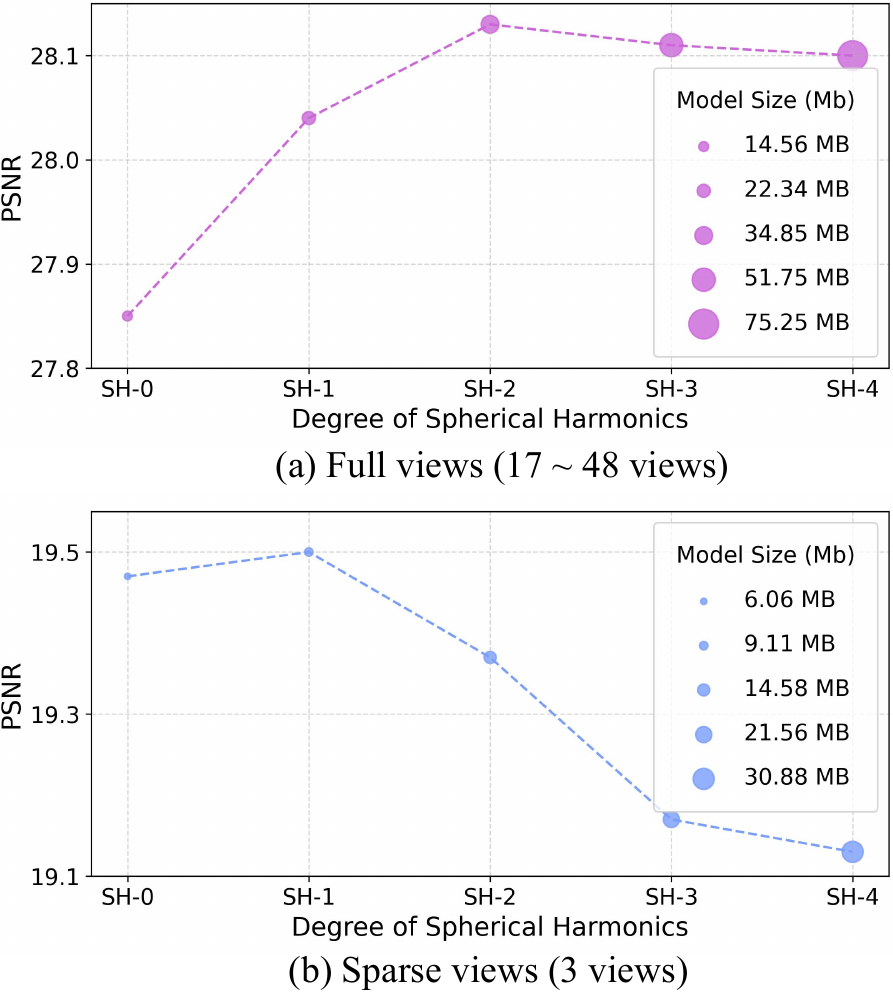} 
\caption{\textbf{Overfitting caused by high-degree SH}. On the LLFF dataset, (a) under full-view conditions, moderately increasing the degree of spherical harmonics can improve the performance of the 3DGS model. However, (b) under sparse-view settings, using high-degree spherical harmonics leads to performance degradation and a significant increase in model size, indicating that spherical harmonics themselves also constitute a source of overfitting.}
\label{fig:pilot_study_sh}
\vspace{3mm}
\end{figure}

Inspired by regularization techniques in deep learning~\cite{srivastava2014Dropout}, recent works introduce the Dropout mechanism into 3DGS~\cite{xu2025Dropoutgs, park2025dropgaussian}. These methods randomly set the opacity of certain Gaussians to zero during training, aiming to prevent over-reliance on specific Gaussians and thereby regularize the model. However, our analysis reveals critical limitations in these approaches:
(1) \textit{Neighbor compensation effect} (Figure~\ref{fig:pilot_study}). 3DGS leverages numerous overlapping Gaussians to collaboratively render a scene, which often exhibits highly similar opacity and color attributes in local regions. When a single Gaussian is dropped, its rendering contribution is easily compensated by neighboring ones, thereby weakening the intended regularization effect of Dropout.
(2) \textit{Limited attribute utilization} (Figure~\ref{fig:pilot_study_sh}). 
Existing dropout strategies in 3DGS exclusively target the opacity attribute, while neglecting the distinct roles of other attributes in overfitting.

To address these limitations, we propose a novel Dropout regularization strategy termed \methodname{}. 
Unlike previous Dropout strategies~\cite{park2025dropgaussian,xu2025Dropoutgs} that discard isolated Gaussians, \methodname{} randomly selects a set of anchor Gaussians and drops these anchors and their surrounding neighbours within a local spatial region. 
This structured Dropout strategy eliminates entire clusters of correlated Gaussians, creating larger-scale “information voids” that actively disrupt spatial coherence and prevent local compensation.
Consequently, the optimization process is compelled to utilize more long-range contextual information to reconstruct the dropped regions, thereby fostering more robust and generalizable scene representations.

Moreover, existing 3DGS Dropout methods are limited to manipulating opacity without considering the nuanced differences among multiple attributes. 
We identify that SH, particularly higher-degree terms, also cause overfitting (as shown in Figure~\ref{fig:pilot_study_sh}). To this end, we further introduce a Dropout mechanism targeting high-degree SH. 
This strategy provides twofold benefits: it (1) mitigates overfitting to color variations, and (2) promotes a coarse-to-fine SH representation, prioritizing lower-degree SH for scene representation. 
Consequently, the trained model permits post-hoc pruning of high-degree SH to obtain a more compact model without requiring retraining, as shown in Figure~\ref{fig:teaser}.

The main contributions of our work are as follows: 
\begin{itemize}
\item We identify and analyze the limitations of existing Dropout strategies in 3DGS, showcasing how spatial redundancy and high-degree spherical harmonic coefficients weaken their regularization effect.
\item We propose \methodname{}, a novel structured spatial Dropout method that discards clusters of Gaussians, achieving stronger regularization against overfitting.
\item We extend the Dropout strategy to appearance attributes by dropping SH, which simultaneously suppresses overfitting and enables flexible post-hoc model compression.
\item Comprehensive experiments demonstrate that our method significantly outperforms the state-of-the-art Dropout techniques for 3DGS, while incurring negligible additional computational cost. Moreover, it can be used to enhance the quality of multiple 3DGS variants, highlighting its broad applicability and value.
\end{itemize}

%% file: sec/2_related_work.tex
\section{Related Work}

\noindent\textbf{NeRF and 3DGS.}
Neural Radiance Fields (NeRF) implicitly model 3D scenes using neural networks to enable high-quality novel view synthesis~\cite{mildenhall2020nerf,mueller2022instant,barron2021mip,zhang2022ray,atzmon2019controlling,mescheder2019occupancy,michalkiewicz2019implicit,niemeyer2019occupancy,oechsle2019texture,park2019deepsdf,peng2020convolutional,lindell2021autoint,sun2022direct}. However, the training and rendering processes of NeRF are computationally intensive, limiting its applicability in real-time scenarios~\cite{picard2023survey,fang2023pvd,reiser2021kilonerf,garbin2021fastnerf,xu2022point-nerf,wang2024uavenerf,martinbrualla2020nerfw,lee2024sharp,fang2023dn2n}. Subsequent efforts, such as Instant-NGP~\cite{mueller2022instant} and Plenoxels~\cite{fridovich2022plenoxels}, improve efficiency through network architecture optimization or by introducing explicit scene representations. Recently, 3DGS has gained significant attention as an efficient explicit representation~\cite{kerbl2023-3dgs,malarz2023-vdgs,lu2024scaffold,huang20242dgs,jiang2024gaussianshader,gao2023relightable,mihajlovic2025splatfields,jung2024relaxing,yang2024gaussianobject}. It models the 3D scene as a set of Gaussian functions, using attributes such as position, covariance, opacity, and spherical harmonic coefficients for fast and high-fidelity rendering. Follow-up works have further enhanced 3DGS in terms of parameter efficiency and rendering quality~\cite{yu2024mip-gs,chen2025hac,lu2024scaffold,liu2024comp-gs,chen2025mvsplat,huang20242dgs,malarz2023-vdgs,chen2024survey-3dgs,bag20243dgs-comp-survey,das2024NPGs,foroutan2024-NerfInit, niemeyer2024radsplat,lee2024compact-3dgs}.

\vspace{1mm}
\noindent\textbf{NVS under Sparse Views.}
Novel view synthesis under sparse views aims to generate high-quality novel views from limited input images. Prior studies~\cite{wang2023sparsenerf,mihajlovic2025splatfields,barron2022mipnerf-360,jain2021dietnerf,niemeyer2022regnerf,colmapfree3dgs,yang2023freenerf} have noted that the performance of both NeRF and 3DGS methods heavily depends on the dense views. To address this, researchers have proposed various methods to enhance NeRF's generalization under sparse views, such as DietNeRF~\cite{jain2021dietnerf}, RegNeRF~\cite{niemeyer2022regnerf}, and FreeNeRF~\cite{yang2023freenerf}, which employ diverse regularization loss strategies to mitigate overfitting. For sparse-view optimization in 3DGS, FSGS~\cite{zhu2024fsgs} introduces a Proximity-guided Gaussian Unpooling strategy, CoR-GS~\cite{zhang2024corgs} uses mutual constraints between two distinct 3DGS models, InstantSplat~\cite{fan2024instantsplat} jointly optimizes poses, while DNGaussian~\cite{li2024dngaussian} and  NexusGS~\cite{zheng2025nexusgs} introduce the depth constraint to enhance the performance of the 3DGS model.

\vspace{1mm}
\noindent\textbf{Dropout Technique in 3DGS.}
In this work, we focus on leveraging the Dropout technique to improve the performance of 3DGS under sparse-view conditions. The most closely related works are DropoutGS~\cite{xu2025Dropoutgs} and DropGaussian~\cite{park2025dropgaussian}, both of which propose mitigating overfitting in 3DGS by randomly discarding Gaussians during training.
However, the high correlation among neighboring Gaussians limits the effectiveness of their Dropout strategies. In contrast, our \methodname{} method significantly strengthens regularization by discarding clusters of neighboring Gaussians and applying Dropout to spherical harmonic coefficients. Our approach achieves a flexible trade-off between performance and model size, offering a novel solution for the sparse-view NVS task.

%% file: sec/3_method.tex
\begin{figure*}[t]
\centering
\includegraphics[width=1. \textwidth]{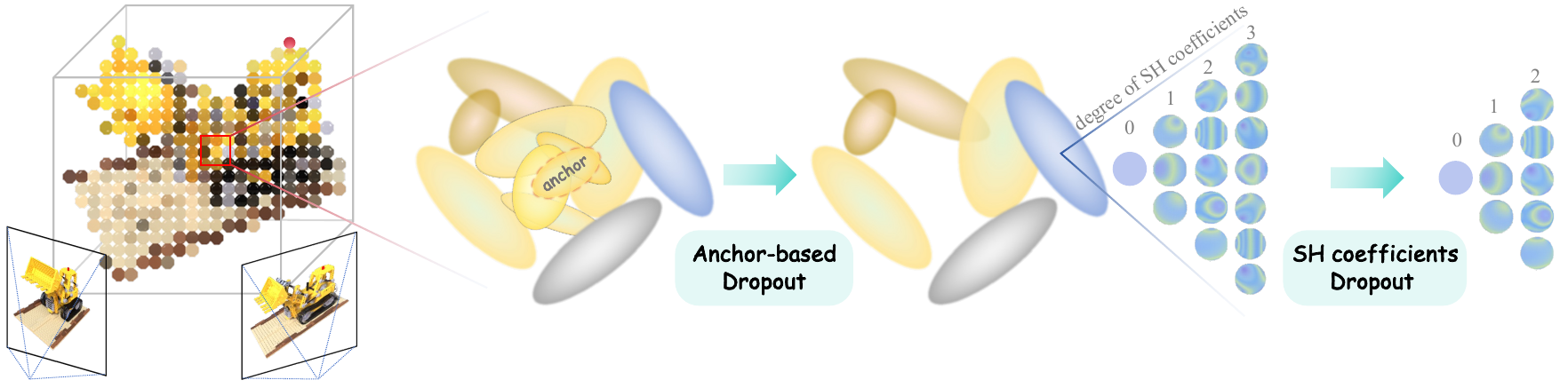} 
\caption{\textbf{Overview of \methodname{}}. During the training phase, we randomly select a set of anchor Gaussians and drop these anchors and their surrounding neighbours within a local spatial region, thereby reducing the likelihood of neighboring Gaussians compensating for one another and enhancing the regularization effect. Simultaneously, high-degree SH coefficients are randomly dropped to further strengthen regularization and yield a more compact model representation.}
\label{fig:method}
\end{figure*}

\section{Method}

\subsection{Preliminaries}
3DGS represents a 3D scene using a large number of explicit Gaussians. Each Gaussian $G_i$ is parameterized by: position $\mu_i \in \mathbb{R}^3$, covariance $\Sigma_i \in \mathbb{R}^{3 \times 3}$, color $c_i$ (represented by a set of spherical harmonics coefficients), and opacity $\alpha_i \in [0, 1]$. For efficient optimization, the covariance $\Sigma_i$ is typically factorized into a scaling vector $s_i \in \mathbb{R}^3$ and a rotation quaternion $q_i \in \mathbb{R}^4$.
Given a camera view defined by a view matrix $W$, the 3D Gaussians are projected onto the 2D image plane. The color $C$ of a pixel is computed using $\alpha$-blending over all $N$ depth-sorted Gaussians:

\begin{equation}
C = \sum_{i=1}^{N} c_i \alpha_i \prod_{j=1}^{i-1} (1 - \alpha_j).
\label{eq:gs}
\end{equation}

\subsection{Pilot Study}
To mitigate overfitting under sparse view settings, prior works propose randomly setting the opacity $\alpha_i$ of Gaussians to zero during training. However, the expressive nature of 3DGS exhibits significant local redundancy. A visible surface is typically represented by hundreds or thousands of overlapping Gaussians. When a single Gaussian is removed, as per Eq.~(\ref{eq:gs}), its contribution to pixel color is compensated by the increased contributions of neighboring Gaussians. Consequently, the change in pixel color $ \Delta C $ is negligible, resulting in weak gradient signals during backpropagation that fail to impose effective regularization on the model’s geometry and appearance learning. 

As illustrated in Figure~\ref{fig:pilot_study}, spatial correlation among Gaussians is inversely related to their distance. Removing a single Gaussian within the view frustum results in minimal performance degradation and weak gradients. In contrast, discarding a cluster of 10 neighboring Gaussians produces significant performance changes and stronger gradient signals. This suggests that Gaussians exhibit strong spatial complementarity. While this facilitates cooperative rendering, it also diminishes the effectiveness of naive Dropout.

Moreover, existing methods focus solely on dropping opacity values, overlooking the regularization potential of other crucial attributes, such as the spherical harmonic coefficients that encode appearance.
As illustrated in Figure~\ref{fig:pilot_study_sh}, while using high-degree SH improves performance under full-view conditions, it leads to overfitting in sparse-view settings. Appropriately reducing the degree of SH not only enhances 3DGS performance in sparse-view scenarios but also significantly reduces the model size.

\begin{figure*}[t]
\centering
\includegraphics[width=1.\textwidth]{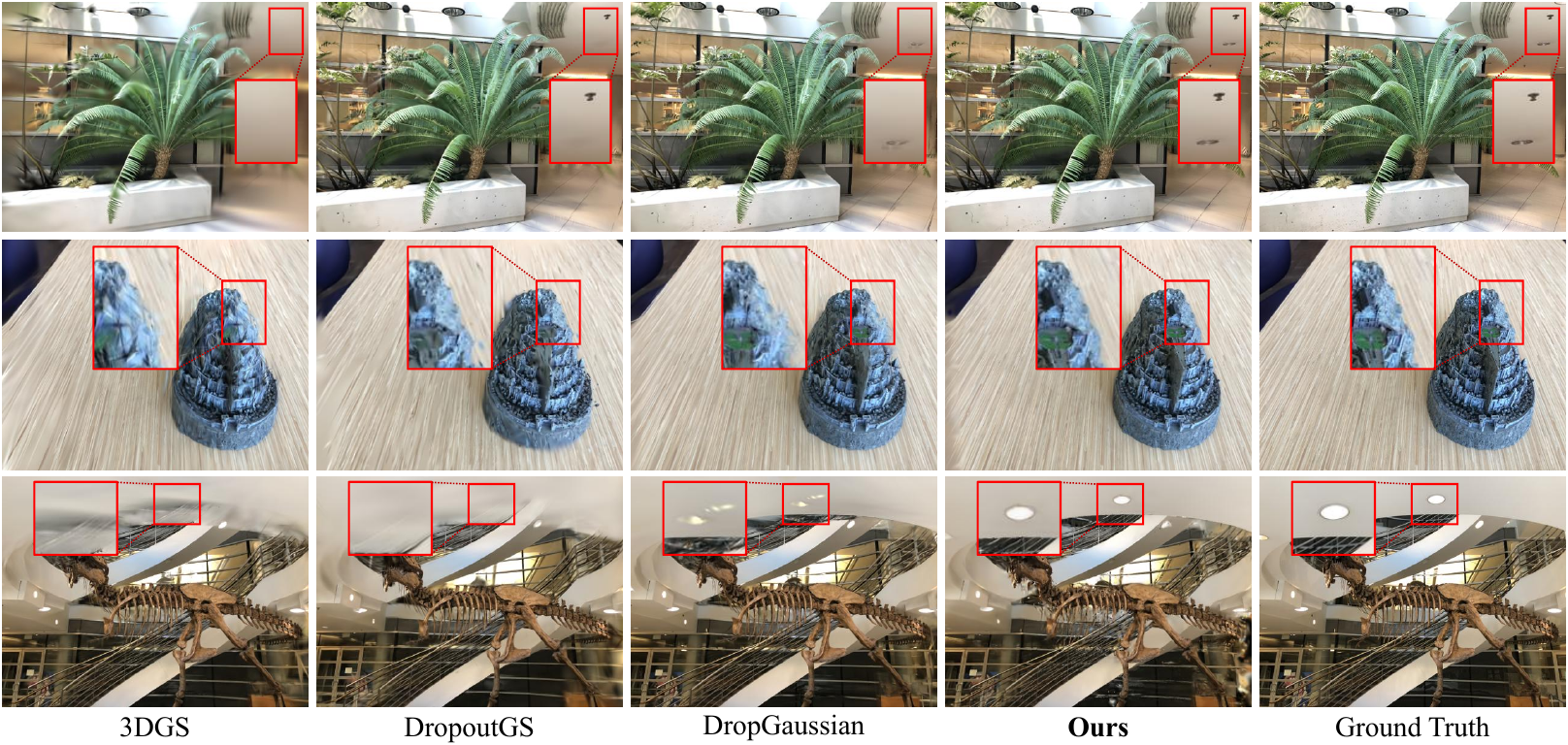} 
 \caption{\textbf{Qualitative comparison on the LLFF dataset (3 views).}}
\label{fig:compare_llff}
\end{figure*}

\begin{table*}[t]
  \centering
  \caption{\textbf{Quantitative comparison on the LLFF dataset (3, 6 and 9 views).}}
  \resizebox{1.\linewidth}{!}{
    \input{tabs/compare_llff}
    }
  \label{tab:compare_llff}%
\end{table*}

\subsection{Anchor-based Dropout}

To overcome the above limitations, we propose \methodname{} method, as illustrated in Figure~\ref{fig:method}, which drops groups of spatially related Gaussians rather than individual, isolated ones. The core idea is to remove entire regions of information, thereby forcing the model to rely on broader contextual cues for reconstruction and encouraging the learning of non-local, structural scene features. This process is executed during training with the following steps:

\noindent \textit{1. Anchor Selection:} We randomly select a subset of anchor Gaussians $ \mathcal{A} \subset \mathcal{G} $ from the set of all $ N $ Gaussians $ \mathcal{G} = \{G_1, \dots, G_N\} $, based on a sampling ratio $ p_a $. 

\noindent \textit{2. Neighborhood Construction:} For each anchor Gaussian in $\mathcal{A}$, we identify its $k$ nearest neighboring Gaussians in Euclidean space based on the distance between their centers. 

\noindent \textit{3. Structured Dropout:} All Gaussians that belong to any anchor or its neighborhood are collected into a Dropout set $\mathcal{D}$.
We then define a binary mask vector $M \in \{0,1\}^N$ for all Gaussians. For each $G_i$, its mask value $m_i$ is:
\begin{equation}
    m_i = \begin{cases} 0 & \text{if } G_i \in \mathcal{D} \\ 1 & \text{otherwise} \end{cases}.
    \label{eq:gs_mask}
\end{equation}
During training, the original opacity $\alpha_i$ of each Gaussian is modulated by its mask to obtain the effective opacity $\hat{\alpha}_i$:
\begin{equation}
    \hat{\alpha}_i = \alpha_i \cdot m_i.
    \label{eq:new_opacity}
\end{equation}
This modulated opacity $\hat{\alpha}_i$ is then used in the $\alpha$-blending process in Eq.~(\ref{eq:gs}), effectively creating “information voids” in the 3D scene and enforcing stronger regularization.

\subsection{Spherical Harmonics Dropout}
 Existing Dropout methods focus solely on the presence or absence of Gaussians according to opacity, overlooking the distinct characteristics of other attributes in Dropout. 
We observe that high-degree SH coefficients are also prone to overfitting. To address this, we propose a Dropout strategy targeting these coefficients.
Specifically, the color $\mathbf{c}$ of a Gaussian is represented by SH coefficients up to degree $L$:
\begin{equation}
\mathbf{c} = [\mathbf{c}^{(0)}, \mathbf{c}^{(1)}, \dots, \mathbf{c}^{(L)}],
\end{equation}
where $\mathbf{c}^{(l)}$ denotes the SH coefficients of degree $l$, comprising $ 2l+1 $ values per RGB channel. 
During training, we randomly select a subset of Gaussians with probability $ p_{sh} $ and set a maximum retained degree $ l_{\text{max}} $, then discard all SH coefficients of degree higher than $ l_{\text{max}} $.

\begin{equation}
\tilde{\mathbf{c}} = [\mathbf{c}^{(0)}, \dots, \mathbf{c}^{(l_\text{max})}, \mathbf{0}, \dots, \mathbf{0}].
\end{equation}
This Dropout forces the model to prioritize low-degree SH coefficients for basic appearance. As training progresses, $l_\text{max}$ gradually increases to allow more high-frequency detail. After training, higher-degree coefficients become optional, allowing users to discard them to obtain a smaller and faster model without retraining. This enables flexible trade-offs between performance and efficiency.

\subsection{Loss Function}
Our proposed strategies can be seamlessly integrated into the training pipeline of existing 3DGS frameworks without modifying their objective functions. We adopt the standard loss that minimizes the difference between the rendered image $\hat{C}$ and the ground truth $C_{gt}$ using a combination of L1 Norm and SSIM losses~\cite{kerbl2023-3dgs}:
\begin{equation}
    \mathcal{L} = \mathcal{L}_{\text{L1}}(\hat{C}, C_{gt}) + \lambda \mathcal{L}_{\text{SSIM}}(\hat{C}, C_{gt}),
    \label{eq:loss}
\end{equation}
where $\lambda$ is a balancing weight that controls the relative contribution of the L1 and SSIM terms.

%% file: tabs/compare_llff.tex
\begin{tabular}{c|c|ccc|ccc|ccc}
\toprule
      & \multirow{2}[4]{*}{Methods} & \multicolumn{3}{c|}{3-view} & \multicolumn{3}{c|}{6-view} & \multicolumn{3}{c}{9-view} \\
\cmidrule{3-11}      &       & PSNR↑ & SSIM↑ & LPIPS↓ & PSNR↑ & SSIM↑ & LPIPS↓ & PSNR↑ & SSIM↑ & LPIPS↓ \\
\midrule
\multirow{4}[2]{*}{NeRF-based} & Mip-NeRF~\cite{barron2022mipnerf-360} & 16.11 & 0.401 & 0.460  & 22.91 & 0.756 & 0.213 & 24.88 & 0.826 & 0.170  \\
      & DietNeRF~\cite{jain2021dietnerf} & 14.94 & 0.370  & 0.496 & 21.75 & 0.717 & 0.248 & 24.28 & 0.801 & 0.183 \\
      & RegNeRF~\cite{niemeyer2022regnerf} & 19.08 & 0.587 & 0.336 & 23.10  & 0.760  & 0.206 & 24.86 & 0.820  & 0.161 \\
      & FreeNeRF~\cite{yang2023freenerf} & 19.63 & 0.612 & 0.308 & 23.73 & 0.779 & 0.195 & 25.13 & 0.827 & 0.160  \\
\midrule
\multirow{7}[4]{*}{3DGS-based} & 3DGS~\cite{kerbl2023-3dgs}  & 19.17 & 0.646 & 0.268 & 23.74  & 0.807  & 0.162 & 25.44 & 0.860  & 0.096 \\
      & DNGaussian~\cite{li2024dngaussian} & 19.12 & 0.591 & 0.294 & 22.18 & 0.755 & 0.198 & 23.17 & 0.788 & 0.180  \\
      & FSGS~\cite{zhu2024fsgs}  & 20.43 & 0.682 & 0.248 & 24.09 & 0.823 & 0.145 & 25.31 & 0.860  & 0.122 \\
      & CoR-GS~\cite{zhang2024corgs} & \underline{20.36} & \underline{0.710} & 0.202 & 24.34 & \underline{0.831} & \underline{0.122} & \underline{25.94} & \underline{0.872} & \textbf{0.088} \\
\cmidrule{2-11}      & DropoutGS~\cite{xu2025Dropoutgs} & 19.39 & 0.632 & 0.279 & 24.02 & 0.816 & 0.144 & 25.13 & 0.869 & 0.099 \\
      & DropGaussian~\cite{park2025dropgaussian} & 20.33 & 0.709 & \underline{0.201} & \underline{24.58} & 0.830  & 0.125  & 25.85 & 0.864 & \underline{0.093} \\
      & Ours  & \textbf{20.68} & \textbf{0.724} & \textbf{0.194} & \textbf{24.76 } & \textbf{0.837} & \textbf{0.116} & \textbf{26.24 } & \textbf{0.875} & \textbf{0.088} \\
\bottomrule
\end{tabular}%

%% file: sec/4_exp.tex
\begin{figure*}[t]
\centering
\includegraphics[width=1.\textwidth]{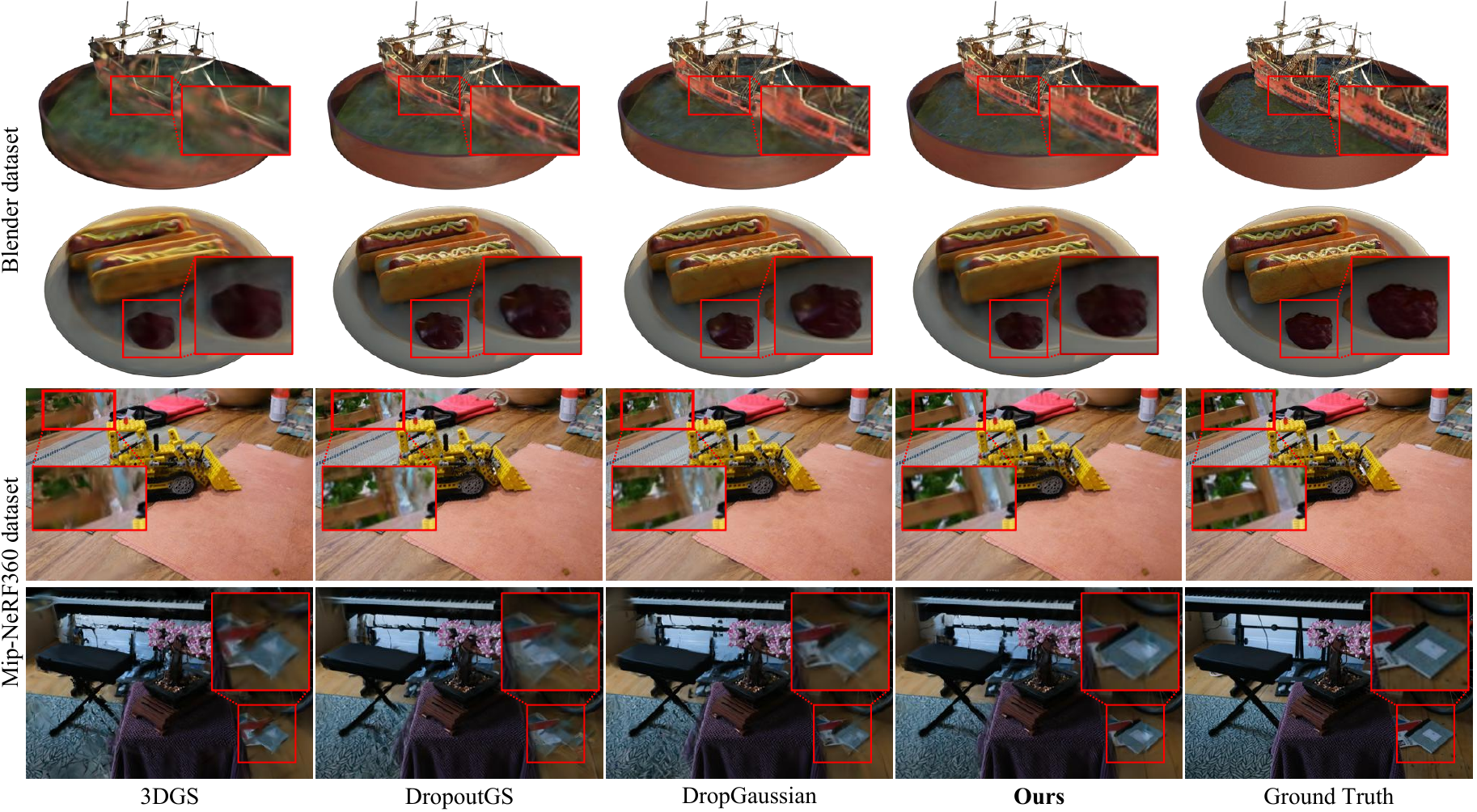} 
\caption{\textbf{Qualitative comparisons on the Blender (8 views) and MipNeRF-360 (12 views) datasets.}}
\label{fig:compare_mip_blender}
\end{figure*}

\section{Experiments}

\subsection{Implementation Details}

\noindent\textbf{Datasets.}  
Our experiments are conducted on three standard datasets: two real-world datasets of LLFF~\cite{mildenhall2019llff-data} and MipNeRF-360~\cite{barron2022mipnerf-360}, and one synthetic dataset of Blender~\cite{mildenhall2020nerf}. We follow previous works~\cite{wang2023sparsenerf,park2025dropgaussian,yang2023freenerf} to split the datasets into training and testing subsets with varying numbers of input views.

\vspace{1mm}
\noindent\textbf{Metrics.}  
To comprehensively evaluate rendering quality, we adopt three widely used metrics: Peak Signal-to-Noise Ratio (PSNR, higher is better), Structural Similarity Index~\cite{wang2004image-ssim} (SSIM, higher is better), and Learned Perceptual Image Patch Similarity~\cite{zhang2018-lpips} (LPIPS, lower is better).

\vspace{1mm}
\noindent\textbf{Baselines.}  
We compare our method against several state-of-the-art approaches, including NeRF-based sparse-view methods (Mip-NeRF~\cite{barron2022mipnerf-360}, RegNeRF~\cite{niemeyer2022regnerf}, DietNeRF~\cite{jain2021dietnerf}, FreeNeRF~\cite{yang2023freenerf}) and 3DGS-based methods (3DGS~\cite{kerbl2023-3dgs}, DNGaussian~\cite{li2024dngaussian}, FSGS~\cite{zhu2024fsgs}, CoR-GS~\cite{zhang2024corgs}). We particularly focus on comparing with two Dropout-based methods, DropGaussian~\cite{park2025dropgaussian} and DropoutGS~\cite{xu2025Dropoutgs}.

\vspace{1mm}
\noindent\textbf{Training Details.}  
All experiments are implemented using the PyTorch framework and conducted on a single NVIDIA H100 GPU. We maintain the same optimizer settings and learning rate strategy as the original 3DGS, training for 10,000 iterations across all datasets. The key hyperparameters are set as follows: the anchor sampling rate $ p_a $ linearly increases from 0 to 0.02 based on the value of iterations; The number of nearest neighboring Gaussians $ k $ is set to 10; the Dropout rate for SH coefficients $ p_{sh} $ is 0.2, with $ l_{\text{max}} $ set to 0, 1, and 2 at 2,000, 4,000, and 6,000 iterations, respectively. For methods involving randomness, including ours, CoR-GS, and the Dropout-based baselines, we report the mean metrics over 3 independent runs. 

For more experimental results, please refer to the supplementary materials.

\begin{table}[t]
  \centering
  \caption{\textbf{Quantitative comparison on the MipNeRF-360 (12 views) and Blender (8 views) datasets}. SH\textsubscript{\textit{n}} denotes retaining only the first \textit{n} degrees of SH coefficients during inference.}
  \resizebox{1.\linewidth}{!}{
    \input{tabs/compare_mip_blender}
    }
\label{tab:compare_mip_blender}
\end{table}

\begin{figure*}[t]
\centering
\caption{\textbf{Compatibility to different 3DGS variants on the LLFF dataset (3 views)}. The proposed \methodname{} can be easily integrated into other 3DGS-based methods and improve their performance.}
\includegraphics[width=1.\textwidth]{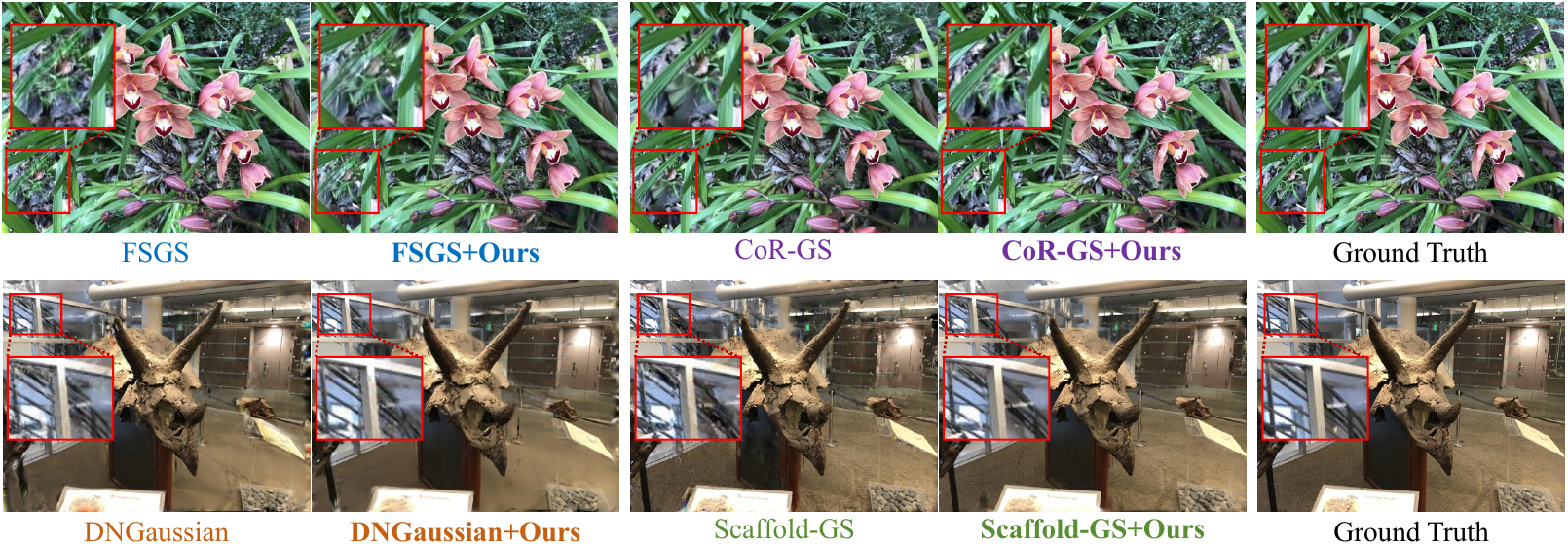} 
\label{fig:compatibility}
\end{figure*}

\subsection{Comparison Results}

\noindent\textbf{Quantitative Comparison.}  
We first perform quantitative evaluations on the LLFF, MipNeRF-360 and Blender datasets under varying sparse view conditions, with results presented in Tables~\ref{tab:compare_llff} and \ref{tab:compare_mip_blender}. 
Our method consistently outperforms all baselines. On the LLFF dataset, particularly in the extremely sparse 3-view setting, our method significantly surpasses other Dropout-based approaches that discard individual Gaussians. This demonstrates the effectiveness of the proposed anchor-based Dropout strategy. As the number of views increases, our approach continues to significantly enhance 3DGS performance. 
Similar performance is observed on the MipNeRF-360 and Blender datasets, confirming that our regularization technique robustly addresses the core challenges of 3DGS under sparse-view conditions.

\vspace{1mm}
\noindent\textbf{Qualitative Comparison.}  
As illustrated in Figures~\ref{fig:compare_llff} and~\ref{fig:compare_mip_blender}, we provide qualitative comparisons across various scenes, illustrating the superiority of our method. 
By removing entire regions, our method simulates occlusions and missing information, prompting Gaussians to store more complete scene context. This leads to more complete and consistent novel view synthesis, as evidenced in Figure~\ref{fig:compare_llff}, where our method preserves more structural detail compared to baselines, which suffer from distortion artifacts and information loss.
Furthermore, baseline methods produce numerous Gaussian-shaped artifacts in Figure~\ref{fig:compare_mip_blender}, especially around object boundaries and in the background. In contrast, our method effectively suppresses these issues by discouraging the model from overfitting complex surfaces with a small number of high-opacity Gaussians, resulting in smoother and more natural geometric structures. 

\subsection{Compression of Model Size} 
A unique aspect of our method is the Dropout of SH coefficients, which enables the scene appearance to be stored incrementally from low-degree to high-degree terms. 
At inference time, the trained model can be truncated by removing high-degree SH coefficients. As shown in Tables~\ref{tab:compare_mip_blender}, even a model retaining only the zeroth-degree SH coefficients outperforms the vanilla 3DGS, while requiring only 25\% of the parameters. This provides a flexible way to balance model size and performance.

\begin{table}[t]
  \centering
  \caption{\textbf{Compatibility to 3DGS variants on the LLFF dataset (3 views)}. The proposed \methodname{} can be easily integrated into various 3DGS-based methods and improve their performance.}
  \resizebox{0.8\linewidth}{!}{
    \input{tabs/compatibility}
    }
\label{tab:compatibility}
\end{table}

\subsection{Compatibility} 
The proposed Dropout framework is lightweight and modular, making it easy to plug into other 3DGS variants. As shown in Table~\ref{tab:compatibility} and Figure~\ref{fig:compatibility}, applying our framework to the FSGS~\cite{zhu2024fsgs}, Cor-GS~\cite{zhang2024corgs}, DNGaussian~\cite{li2024dngaussian}, and Scaffold-GS~\cite{lu2024scaffold} significantly improves their performance under sparse view conditions. This suggests that our method is broadly applicable and can enhance the robustness of other 3DGS frameworks in view-scarce scenarios.

\begin{table}[t]
\centering
\caption{\textbf{Comparison of training time on different datasets.} We achieve a substantial performance improvement with a negligible increase in computational time.}
\resizebox{0.8\linewidth}{!}{
\begin{tabular}{c|lccl}
\toprule
\multirow{2}[4]{*}{Method} & PSNR↑ & SSIM↑ & LPIPS↓ & Total Time↓ \\
\cmidrule{2-5}      & \multicolumn{4}{c}{LLFF dataset (3 views)} \\
\midrule
3DGS  & 19.17 & 0.646 & 0.268 & \textbf{741.6s} \\
Ours  & \textbf{20.68} & \textbf{0.724} & \textbf{0.194} & 760.2s \\
\midrule
      & \multicolumn{4}{c}{Blender dataset (8 views)} \\
\cmidrule{2-5}3DGS  & 22.13 & 0.855 & 0.132 & \textbf{863.3s} \\
Ours  & \textbf{25.50} & \textbf{0.891} & \textbf{0.088} & 887.7s \\
\midrule
      & \multicolumn{4}{c}{MipNeRF-360 dataset (12 views)} \\
\cmidrule{2-5}3DGS  & 18.58 & 0.526 & 0.419 & \textbf{1083.2s} \\
Ours  & \textbf{19.93} & \textbf{0.576} & \textbf{0.362} & 1114.8s \\
\bottomrule
\end{tabular}%
}
\label{tab:time_comparison}
\end{table}

\begin{table*}[t]
  \centering
  \caption{\textbf{Hyperparameter sensitivity of the anchor selection probability $p_a$, number of neighbors $k$, and SH Dropout probability $p_{sh}$}. The evaluation is conducted on the LLFF dataset (3 views).}
  \resizebox{0.8\linewidth}{!}{
\begin{tabular}{c|ccc|c|ccc|c|ccc}
\toprule
$p_a$ & PSNR↑ & SSIM↑ & LPIPS↓ & $k$   & PSNR↑ & SSIM↑ & LPIPS↓ & $p_{sh}$ & PSNR↑ & SSIM↑ & LPIPS↓ \\
\midrule
0.005 & 20.25 & 0.705 & 0.211 & 1     & 20.12 & 0.701 & 0.216 & 0.1   & 20.56 & 0.721 & 0.198 \\
0.01  & 20.50  & 0.718  & \textbf{0.194} & 5     & 20.39 & 0.719 & 0.200  & 0.2   & \textbf{20.68} & 0.724 & \textbf{0.194} \\
0.02  & \textbf{20.68} & \textbf{0.724} & \textbf{0.194} & 10    & \textbf{20.68} & \textbf{0.724} & \textbf{0.194} & 0.3   & 20.65 & \textbf{0.725} & 0.197 \\
0.03  & 20.32 & 0.710  & 0.202 & 15    & 20.43 & 0.721 & 0.198 & 0.4   & 20.44 & 0.713 & 0.202 \\
0.04  & 19.97 & 0.665 & 0.257 & 20    & 20.07 & 0.706 & 0.224 & 0.5   & 20.49 & 0.711 & 0.208 \\
\bottomrule
\end{tabular}%
    }
  \label{tab:sensitivity}
\end{table*}

\begin{table}[t]
  \centering
  \caption{\textbf{Ablation results of the designed components on the LLFF datasets (3 views).}}
  \resizebox{0.9\linewidth}{!}{
    \input{tabs/ablation_component}
    }
\label{tab:ablation_component}
\end{table}

\begin{table}[ht]
\centering
\caption{\textbf{Ablation results of different SH Dropout strategies on the Blender dataset (8 views).}}
\resizebox{0.88\linewidth}{!}{
\begin{tabular}{l|ccc}
\toprule
      & PSNR↑ & SSIM↑ & LPIPS↓ \\
\midrule
Drop SH Randomly & 25.12 & 0.885 & 0.097  \\
Drop SH by Degree (Ours) & \textbf{25.50 } & \textbf{0.891} & \textbf{0.088} \\
\bottomrule
\end{tabular}
}
\label{tab:sh_ablation}
\end{table}

\subsection{Training Efficiency}
Our method introduces an additional computational step during training, i.e., nearest neighbor search for anchor Gaussians. To accelerate this process, we implement it efficiently using CUDA on GPUs~\cite{Garcia2010knn-cuda}.
We run our method and the vanilla 3DGS on the same NVIDIA H100 GPU for 10,000 iterations to compare their total training time on different datasets.
As shown in Table~\ref{tab:time_comparison}, our method incurs only a modest increase in training time (less than 2.8\% compared to 3DGS). Given the substantial improvement in rendering quality (an average PSNR gain of 2 dB), this marginal computational cost is well justified.

\subsection{Ablation Study}
\noindent\textbf{Effectiveness of each dropout component}. We conduct ablation studies in Table~\ref{tab:ablation_component}, revealing the following:  
(1) Removing the ``Drop Anchor" strategy leads to significant performance declines across all metrics, confirming the efficacy of our spatial Anchor-based Dropout in breaking local redundancy for regularization.
(2) Omitting the ``Drop SH" module also results in performance degradation, demonstrating that regularizing appearance attributes is another effective approach to mitigating overfitting.
The full model achieves the best performance, indicating that Drop Anchor and Drop SH are complementary and jointly enhance the model’s overall performance.

\vspace{1mm}
\noindent\textbf{Drop SH by Degree \textit{vs} Drop SH Randomly.}
In \methodname{}, we use a ``Drop SH by Degree" strategy for discarding high-degree SH coefficients during training. 
We compare it against a more direct baseline: ``Drop SH Randomly". In this alternative strategy, we do not operate on entire degrees but instead randomly set individual SH coefficients to zero with the same probability.
The results in Table~\ref{tab:sh_ablation} demonstrate that our ``Drop SH by Degree" strategy significantly outperforms randomly dropping individual SH coefficients.

\vspace{1mm}
\noindent\textbf{Hyperparameter Sensitivity Analysis}.
We conduct a sensitivity analysis on three key hyperparameters: (1) anchor selection probability $p_a$ (default: 0.02), (2) number of neighbors $k$ (default: 10), and (3) SH Dropout probability $p_{sh}$ (default: 0.2). Experiments are performed on the LLFF dataset (3 views), with results in Table~\ref{tab:sensitivity}.
The parameter $p_a$ controls the proportion of Gaussians selected as anchors. A small $p_a$ gives insufficient regularization. Conversely, an excessively large $p_a$ drops too many Gaussians, which can disrupt the scene's geometric structure and degrade performance.
The Number of Neighbors $k$ determines the size of the ``information voids" around each anchor.
A small $k$ creates voids that are too small and easily compensated for, weakening the regularization effect; too large $k$ removes critical local structures and hurts performance.
The probability $p_{sh}$ controls the dropout of high-degree SH coefficients to prevent overfitting to fine color details. A lower value of $p_{sh}$ provides insufficient regularization, while a higher value overly penalizes the SH coefficients, degrading performance.

%% file: tabs/compare_mip_blender.tex
\begin{tabular}{l|cc|cc}
\toprule
      & \multicolumn{2}{c|}{MipNeRF-360} & \multicolumn{2}{c}{Blender} \\
\midrule
Methods & PSNR↑ & Size (Mb)↓ & PSNR↑ & Size (Mb)↓ \\
\midrule
3DGS~\cite{kerbl2023-3dgs}  & 18.58 & 143.4 & 22.13 & 6.5 \\
DNGaussian~\cite{li2024dngaussian} & 18.84 & 80.7  & 24.31 & 5.5 \\
FSGS~\cite{zhu2024fsgs}  & 18.80  & 172.1 & 24.64 & 8.4 \\
CoR-GS~\cite{zhang2024corgs} & 19.42 & 117.4 & 24.31 & 6.3 \\
DropoutGS~\cite{xu2025Dropoutgs} & 19.17 & 68.4  & 24.79 & 5.2 \\
DropGaussian~\cite{park2025dropgaussian} & 19.66 & 120.7 & 25.17 & 6.0  \\
\midrule
Ours-SH\textsubscript{0} & 19.71 & \textbf{33.8} & 25.04 & \textbf{1.7} \\
Ours-SH\textsubscript{1} & 19.86 & \underline{51.8} & 25.34 & \underline{2.6} \\
Ours-SH\textsubscript{2} & \textbf{19.95} & 81.1  & \underline{25.47} & 4.1 \\
Ours-SH\textsubscript{3} & \underline{19.93} & 122.6 & \textbf{25.50 } & 6.2 \\
\bottomrule
\end{tabular}%

%% file: tabs/compatibility.tex
\begin{tabular}{l|ccc}
\toprule
 Methods     & PSNR↑ & SSIM↑ & LPIPS↓ \\
\midrule
FSGS~\cite{zhu2024fsgs}  & 20.43 & 0.682 & 0.248 \\
FSGS+Ours & \textbf{20.72} & \textbf{0.713} & \textbf{0.205} \\
\midrule
CoR-GS~\cite{zhang2024corgs} & 20.36 & 0.710 & \textbf{0.202} \\
CoR-GS+Ours & \textbf{20.74} & \textbf{0.718} & 0.203 \\
\midrule
DNGaussian~\cite{li2024dngaussian} & 19.12 & 0.591 & 0.294 \\
DNGaussian+Ours & \textbf{19.71} & \textbf{0.639} & \textbf{0.268} \\
\midrule
Scaffold-GS~\cite{lu2024scaffold} & 18.62 & 0.643 & 0.258 \\
Scaffold-GS+Ours & \textbf{19.84} & \textbf{0.679} & \textbf{0.212} \\
\bottomrule
\end{tabular}%

%% file: tabs/ablation_component.tex
\begin{tabular}{cc|ccc}
\toprule
\multicolumn{2}{c|}{Settings} & \multirow{2}[4]{*}{PSNR↑} & \multirow{2}[4]{*}{SSIM↑} & \multirow{2}[4]{*}{LPIPS↓} \\
\cmidrule{1-2}Drop Anchor & Drop SH &       &       &  \\
\midrule
$\times$ & $\times$ & 19.17 & 0.646 & 0.268 \\
$\checkmark$ & $\times$ & 20.47 & 0.713 & 0.200  \\
$\times$ & $\checkmark$ & 19.59 & 0.641 & 0.247 \\
$\checkmark$ & $\checkmark$ & \textbf{20.68} & \textbf{0.724} & \textbf{0.194} \\
\bottomrule
\end{tabular}%

%% file: sec/5_conclusion.tex
\section{Conclusions}
In this work, we address the severe overfitting issue of 3DGS under sparse views.
We are the first to identify that the existing methods based on independent random Dropout stem from the spatial redundancy of Gaussians, which enables local compensation effects to undermine the regularization strength. We further observe that high-degree spherical harmonic coefficients contribute to overfitting in 3DGS.
To this end, we propose \methodname{}, a novel Dropout strategy that discards entire spatial regions rather than isolated Gaussians and effectively disrupts local information dependencies, compelling the model to learn more robust global scene representations. 
Furthermore, we extend Dropout to appearance attributes to reduce overfitting and enable post-training scalability. Our method is simple yet effective, significantly enhancing the 3DGS and its variants under limited-view conditions.

%% file: sec/X_suppl.tex
\clearpage
\setcounter{page}{1}
\maketitlesupplementary

We present additional experiments and analyses to further validate the effectiveness and robustness of our proposed \methodname{}. Specifically, we include: (1) extended ablation studies on the spherical harmonics (SH) dropout strategy, (2) visualizations of reconstructed 3D Gaussian scenes, (3) detailed quantitative comparisons on the Mip-NeRF 360 and Blender datasets, and (4) a discussion of potential directions for future work.

\section{Additional Experimental Results}

\begin{table}[ht]
  \centering
  \caption{Effect of with and without Drop SH strategy on the Blender dataset (8 views). The Drop SH strategy encourages the effective parameters to concentrate in lower-degree SH coefficients, thereby mitigating the performance degradation caused by direct SH truncation.}
  \resizebox{0.7\linewidth}{!}{
    \input{tabs/ablation_sh}
    }
\label{tab:ablation_sh}
\vspace{3mm}
\end{table}

\noindent\textbf{Additional Ablation Study.} We further evaluate the impact of the Drop SH strategy on model performance under different levels of SH truncation. On the Blender dataset with 8 input views, we train two sets of models, with and without the Drop SH strategy, and truncate their SH coefficients to varying degrees during inference. The average performance metrics are summarized in Table~\ref{tab:ablation_sh}. As shown, directly truncating high-degree SH coefficients significantly degrades performance when Drop SH is not applied. In contrast, models trained with Drop SH exhibit much smaller performance drops, indicating that, beyond mitigating overfitting, Drop SH effectively concentrates information in lower-degree SH components. This enables users to reduce model size via SH truncation while maintaining high rendering quality.

\begin{figure*}[t]
\centering
\includegraphics[width=0.9\textwidth]{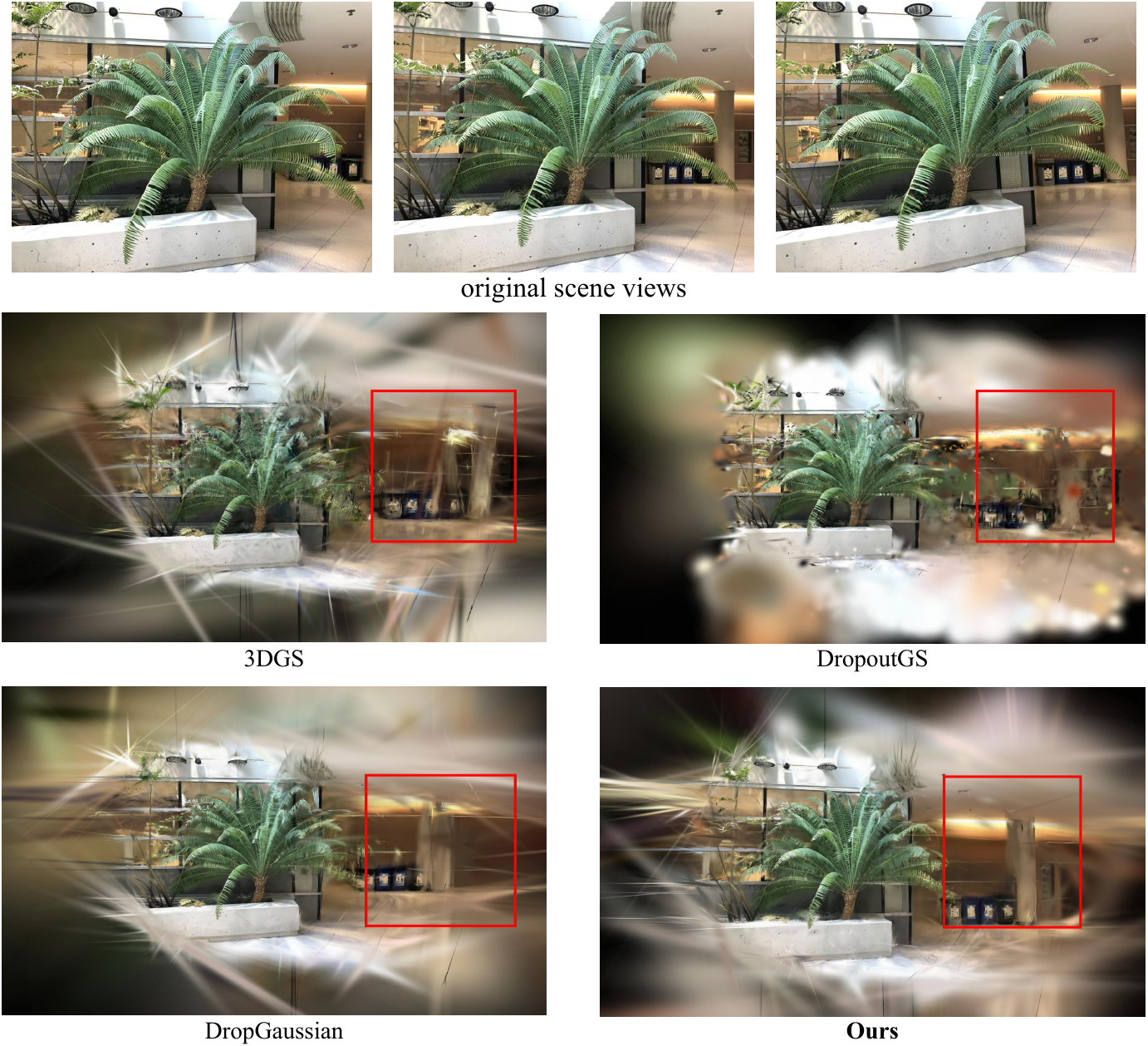} 
 \caption{\textbf{Visualization of the entire reconstruction scene.} Our method produces results that are more complete and natural.}
\label{fig:vis_3d}
\end{figure*}

\vspace{1mm}
\noindent\textbf{Visualization of the Entire Reconstruction Scene.} 
In Figure~\ref{fig:vis_3d}, we present the reconstructed entire scenes. It can be observed that our method produces results that are more complete and natural.
Our method fundamentally mitigates the neighbor compensation effect by randomly selecting a set of “anchors” and simultaneously dropping their spatial neighbors, thereby creating larger information voids. This design encourages remaining Gaussian to rely on a more comprehensive scene context, serving as an effective form of regularization that alleviates overfitting. 
As a result, even when trained under sparse-view conditions, our model develops a more complete and global understanding of the scene’s geometric structure. When rendering from novel views, it can thus reconstruct more coherent and natural structures. 
In contrast, baseline methods tend to over-rely on local information, an approach that may work under dense-view settings but easily leads to overfitting and local artifacts when visual information is limited.

\vspace{1mm}
\noindent\textbf{Additional Comparison Results.}
In Tables~\ref{tab:compare_mip} and~\ref{tab:compare_blender}, we present supplementary quantitative results that complement those in Table~\ref{tab:compare_mip_blender} of the main text. These results consistently demonstrate that our method significantly outperforms existing 3DGS variants tailored for sparse-view settings, delivering both higher rendering quality and more compact model representations.

\section{Discussion on Future Work}
The anchor selection mechanism in our \methodname{} is based on uniform random sampling. However, since the distribution of Gaussians is often non-uniform and their importance can vary across different parts of a 3D scene, exploring more sophisticated anchor selection mechanisms is a valuable direction. For instance, strategies based on gradient magnitude or opacity could be more effective, as these properties often highlight regions that are more critical during optimization. 
Furthermore, \methodname{} selects neighbors for an anchor Gaussian based solely on the Euclidean distance between their positions. This simple spatial proximity metric, however, may not be optimal. Due to the anisotropic nature of Gaussians and the local differences of 3D scenes, physically adjacent Gaussians are not always the most functionally complementary. Therefore, designing more sophisticated neighbor selection schemes presents a promising avenue for future research. Such strategies could incorporate other factors, such as intrinsic Gaussian attributes, local scene characteristics, or even view-dependent information to better identify and regularize clusters of functionally redundant Gaussians.

\begin{table}[t]
  \centering
  \caption{\textbf{Quantitative comparison on the MipNeRF-360 dataset (12 views)}. SH\textsubscript{\textit{n}} denotes retaining only the first \textit{n} degrees of SH coefficients during inference.}
  \resizebox{1.\linewidth}{!}{
    \input{tabs/compare_mip}
    }
\label{tab:compare_mip}%
\end{table}

\begin{table}[t]
  \centering
   \caption{\textbf{Quantitative comparison on the Blender dataset (8 views)}. SH\textsubscript{\textit{n}} denotes retaining only the first \textit{n} degrees of SH coefficients during inference.}
  \resizebox{1.\linewidth}{!}{
    \input{tabs/compare_blender}
    }
\label{tab:compare_blender}%
\end{table}

%% file: tabs/ablation_sh.tex
\begin{tabular}{c|ccc}
\toprule
      & PSNR↑ & SSIM↑ & LPIPS↓ \\
\midrule
w/o - SH\textsubscript{0} & 24.57 & 0.879 & 0.112 \\
w/ - SH\textsubscript{0} & \textbf{25.04} & \textbf{0.883} & \textbf{0.093} \\
\midrule
w/o - SH\textsubscript{1} & 24.96 & 0.885 & 0.096 \\
w/ - SH\textsubscript{1} & \textbf{25.34} & \textbf{0.890 } & \textbf{0.090 } \\
\midrule
w/o - SH\textsubscript{2} & 25.21 & 0.887 & 0.091 \\
w/ - SH\textsubscript{2} & \textbf{25.47} & \textbf{0.891} & \textbf{0.089} \\
\midrule
w/o - SH\textsubscript{3} & 25.36 & 0.890  & 0.090  \\
w/ - SH\textsubscript{3} & \textbf{25.50 } & \textbf{0.891} & \textbf{0.088} \\
\bottomrule
\end{tabular}%

%% file: tabs/compare_mip.tex
\begin{tabular}{l|cccc}
\toprule
Methods & PSNR↑ & SSIM↑ & LPIPS↓ & Size (MB)↓ \\
\midrule
3DGS  & 18.58 & 0.526 & 0.419 & 143.4 \\
DNGaussian & 18.84 & 0.543 & 0.468 & 80.7 \\
FSGS  & 18.80  & 0.531 & 0.418 & 172.1 \\
CoR-GS & 19.42 & 0.556 & 0.418 & 117.4 \\
DropoutGS & 19.17 & 0.558 & 0.386 & 68.4 \\
DropGaussian & 19.66 & 0.569 & 0.374 & 120.7 \\
\midrule
Ours-SH\textsubscript{0} & 19.71 & 0.563 & 0.377 & \textbf{33.8} \\
Ours-SH\textsubscript{1} & 19.86 & \underline{0.571} & 0.371 & \underline{51.8} \\
Ours-SH\textsubscript{2} & \textbf{19.95} & \textbf{0.576} & \underline{0.363} & 81.1 \\
Ours-SH\textsubscript{3} & \underline{19.93} & \textbf{0.576} & \textbf{0.362} & 122.6 \\
\bottomrule
\end{tabular}%

%% file: tabs/compare_blender.tex
\begin{tabular}{c|cccc}
\toprule
Methods & PSNR↑ & SSIM↑ & LPIPS↓ & Size (Mb)↓ \\
\midrule
3DGS  & 22.13 & 0.855 & 0.132  & 6.5 \\
DNGaussian & 24.31 & 0.886 & 0.088 & 5.5 \\
FSGS  & 24.64 & \textbf{0.895} & 0.095 & 8.4 \\
CoR-GS & 24.31 & 0.886 & 0.091 & 6.3 \\
DropoutGS & 24.79 & 0.877 & 0.110  & 5.2 \\
DropGaussian & 25.17 & 0.882 & 0.100  & 6.0  \\
\midrule
Ours-SH\textsubscript{0} & 25.04 & 0.883 & 0.093 & \textbf{1.7} \\
Ours-SH\textsubscript{1} & 25.34 & 0.890  & 0.090  & \underline{2.6} \\
Ours-SH\textsubscript{2} & \underline{25.47} & \underline{0.891} & \underline{0.089} & 4.1 \\
Ours-SH\textsubscript{3} & \textbf{25.50 } & \underline{0.891} & \textbf{0.088} & 6.2 \\
\bottomrule
\end{tabular}%